\RequirePackage{amsmath}
\documentclass[runningheads]{article}
\usepackage{makeidx}
\usepackage{graphicx}
\usepackage{amsmath,amssymb} 
\usepackage{color}

\usepackage{color}
\usepackage{pifont}
\usepackage{multirow}
\usepackage{pgfplots}
\usepackage{pgf,tikz}
\usepackage{mathrsfs}
\usepackage{xspace}

\pgfplotsset{compat=1.13}
\usepackage{url}
\urldef{\mailsa}\path|{follmann, drost, boettger}@mvtec.com|    

\makeatletter
\DeclareRobustCommand\onedot{\futurelet\@let@token\@onedot}
\def\@onedot{\ifx\@let@token.\else.\null\fi\xspace}

\def\eg{\emph{e.g}\onedot} 
 
\def\cf{\emph{c.f}\onedot} 
 \def\vs{\emph{vs}\onedot}
 
\def\etal{\emph{et~al}\onedot}
\makeatother
\newcommand{\figref}[1]{\mbox{Figure~\ref{#1}}}
\newcommand{\tabref}[1]{\mbox{Table~\ref{#1}}}
\newcommand{\secref}[1]{\mbox{Section~\ref{#1}}}

\begin{document}
	\pagestyle{headings}

	\title{Acquire, Augment, Segment \& Enjoy: \\Weakly Supervised Instance Segmentation of Supermarket Products}

	\author{Patrick Follmann\textsuperscript{+*}, Bertram Drost\textsuperscript{+}, and Tobias B\"ottger\textsuperscript{+*}}
	\date{${}^{+}$MVTec Software GmbH, Munich, Germany\\
	${}^{*}$Technical University of Munich (TUM) \\
	\mailsa\\
	\today}

	\maketitle

	\begin{abstract}
		Grocery stores have thousands of products that are usually identified using barcodes with a human in the loop. For automated checkout systems, it is necessary to count and classify the groceries efficiently and robustly. One possibility is to use a deep learning algorithm for instance-aware semantic segmentation. Such methods achieve high accuracies but require a large amount of annotated training data.
    
    We propose a system to generate the training annotations in a weakly supervised manner, drastically reducing the labeling effort. We assume that for each training image, only the object class is known. The system automatically segments the corresponding object from the background. The obtained training data is augmented to simulate variations similar to those seen in real-world setups.
    
    Our experiments show that with appropriate data augmentation, our approach obtains competitive results compared to a fully-supervised baseline, while drastically reducing the amount of manual labeling.
        
\end{abstract}

\begin{figure*}
  \centering
  \includegraphics[width=\textwidth, trim=1cm 15cm 1cm 2.0cm,]{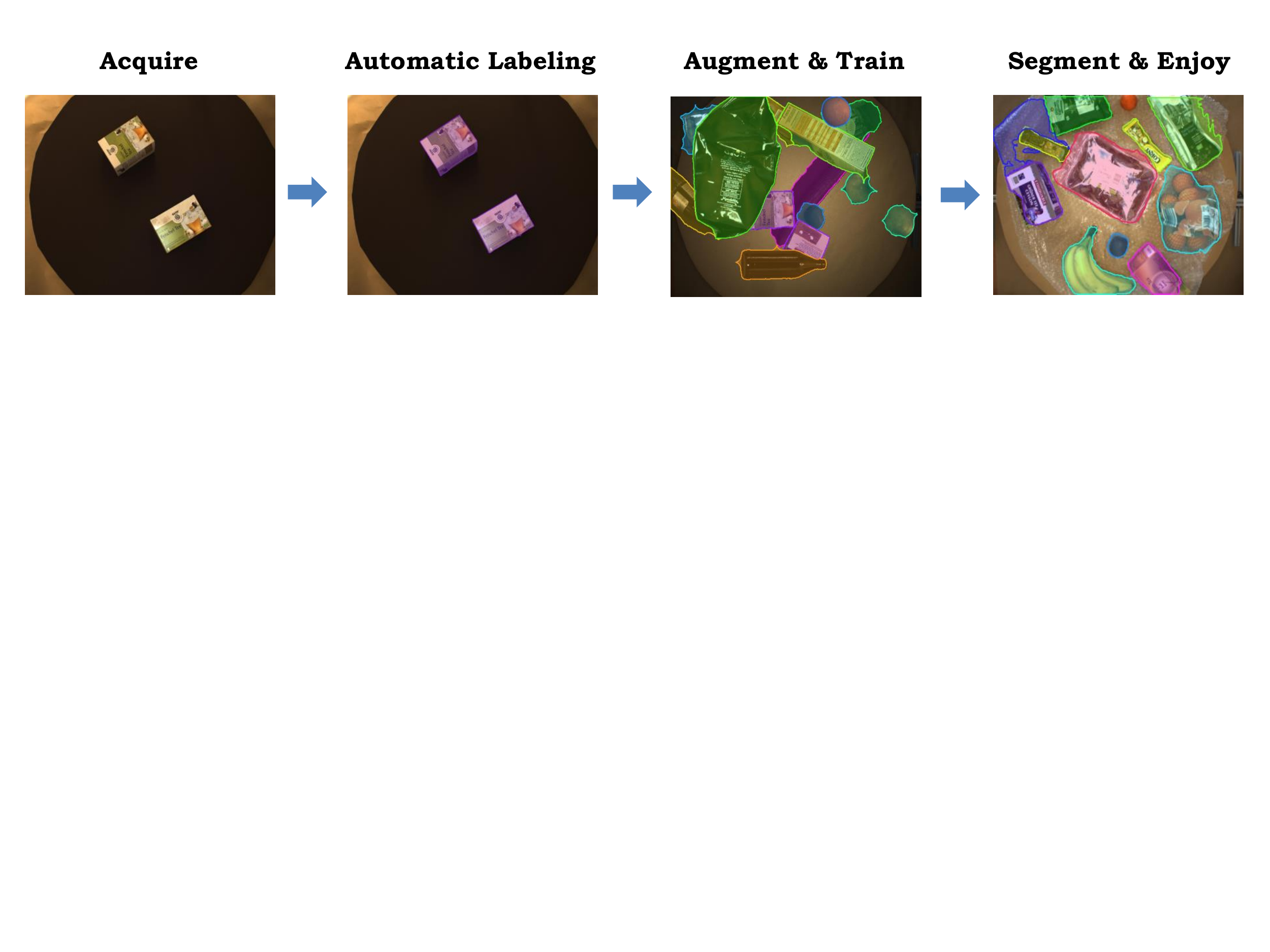}
  \label{fig:eyecatcher}
\end{figure*}


	\section{Introduction}
    
    The classification and localization of objects are important subtasks in many computer vision applications, such as autonomous driving, grasping objects with a robot, or quality control in a production process.
    Recently, end-to-end trainable convolutional neural networks (CNNs) for instance segmentation have been successfully applied in the settings of everyday photography or urban street scenes. This is possible due to the advances in CNN architectures \cite{he_2017_mask} and due to the availability of large-scale datasets, such as ImageNet \cite{Russakovsky_2015_imagenet}, COCO \cite{lin_2014_coco} or Cityscapes \cite{cordts_2016_cityscapes}. For many industrial challenges, the object categories are very specific and the intra- and inter-class variability is rather small. For example, existing automatic checkout systems in supermarkets identify isolated products that are conveyed on a belt through a scanning tunnel \cite{ecrs2018,itab2018}. Even though such systems often provide a semi-controlled environment and know which products may appear, external influences and intra-class variations cannot be completely avoided. Furthermore, the system's efficiency is higher if non-isolated products can be identified as well. To fine-tune a network for such an application, it is crucial to have a large amount of annotated training data. However, the manual annotation of such a dataset is time-consuming and expensive. Hence, it is of great interest to be able to train instance segmentation models with as little labeling effort as possible. 
    
    A recent dataset for this challenge  is D2S \cite{follmann2018d2s}, which contains 21000 images of 60 common supermarket products. The overall objective of that dataset is to realistically model real-world applications such as an automatic checkout, inventory, or warehouse system. It contains very few training images, which are additionally significantly less complex and crowded than the validation and test images. However, the training images are not simple enough to efficiently use weak supervision to generate labels. 
        
    In this work, we present a system that allows to train an instance segmentation model in an industrial setting like D2S with weak supervision. To facilitate this, each of the D2S object categories is captured individually on a turntable. The object regions are labeled automatically using basic image processing techniques. The only manual input is the class of the object on the turntable. This allows to create annotations for instance-aware semantic segmentation of reasonable quality with minimal effort, essentially by only taking a few images of each object category.
%
%
%
    
    In a second step, we assemble complex training scenes using various kinds of data augmentation like the ones proposed in \cite{follmann2018d2s}. Moreover, to address the challenges in the D2S validation and test set, i.e., reflections, background variations or neighboring objects of the same class, we introduce two new data augmentation stages that additionally model lighting variations and occlusion.
%
%
    
%
In our experiments, we thoroughly evaluate the weakly generated annotations against the baseline trained with fully-supervised training data. Due to partially different objects, a different acquisition setup and lighting changes there is a domain-shift to the validation images. Nevertheless, we find that the proposed method allows for an overall detection performance of 68.9\% compared to 80.1\% of a fully-supervised baseline without domain-shift. Hence, it is possible to produce competitive segmentation results with a very simple acquisition setup, virtually no label effort, and suitable data augmentation.
%
	
    \section{Related Work}
            
    \paragraph{\textbf{Weakly supervised instance segmentation.}}
    Solving computer vision tasks with weakly annotated data has become a major topic in recent years. It can significantly reduce the amount of manual labeling effort and thus make solving new tasks feasible. For example, Deselaers \etal \cite{deselaers2012weakly} learn \emph{generic knowledge} about object bounding boxes from a subset of object classes. This is used to support learning new classes without any location annotation. This allows them to train a functional object detector from weakly supervised images only. Similarly, Vezhnevets \etal \cite{vezhnevets2011weakly} attempt to learn a supervised semantic segmentation method from weak labels that can compete with a fully supervised one. Recently, there has been work that attempts to train instance segmentation models from bounding box labels or by assuming only parts of the data is labeled with pixel-precise annotations. For example, the work by Hu \etal \cite{hu2018learning} attempts to train instance segmentation models over a large set of categories, where most instances only have box annotations. They merely assume a small fraction of the instances have mask annotations that have been manually acquired. Khoreva \etal \cite{khoreva_2017_CVPR} train an instance segmentation model by using GrabCut \cite{rother2004grabcut} foreground segmentation on bounding box ground truth labels.   
    
 In contrast to the above works, our weak supervision only assumes the object class of each training image and does not require bounding boxes of the single objects or their pixel-precise annotations. We use basic image processing techniques and a simple acquisition setup to learn competitive instance segmentation models from weak image annotations.
        
    \paragraph{\textbf{Data augmentation.}}
    Since we restrict the training images to objects of a single class on a homogeneous background, it is essential to augment the training data with more complex, artificial, images. Otherwise, state-of-the-art instance segmentation methods fail to generalize to more complex scenes, different backgrounds or varying lighting conditions. This is often the case for industrial applications, where a huge effort is necessary to obtain a large amount of annotated data. Hence, extending the training dataset by data augmentation is common practice \cite{simard2003best}. Augmentation is often restricted to global image transformations such as random crops, translations, rotations, horizontal reflections, or color augmentations \cite{krizhevsky2012imagenet}. However, for instance-level segmentations, it is possible to extend these techniques to generate completely new images. For example, in \cite{follmann2018amodal,li2016amodal,zhu2017semantic}, new artificial training data is generated by randomly sampling objects from the training split of COCO \cite{lin_2014_coco} and pasting them into new training images. However, since the COCO segmentations are coarse and the intraclass variation is extremely high, the augmentation brings limited gain. On the other hand, in the D2S dataset \cite{follmann2018d2s}, it is difficult to obtain reasonable segmentation results without any data augmentation. The training set is designed to mimic the restrictions of industrial applications, which include little training data and potentially much more complex test images than training images.
    
    Analogously to the above works, we perform various types of data augmentation to increase complexity and the amount of the training data. However, we go a step further and address specific weaknesses of the state of the art on D2S by explicitly generating artificial scenes with neighboring and touching objects. To furthermore gain robustness to changing illumination, we also render the artificial scenes under different lighting conditions by exploiting the depth information. 
    
    \section{Weak annotations}
The goal is to generate annotations for the Densely Segmented Supermarket dataset (D2S) \cite{follmann2018d2s} with as little effort as possible. For this, we built an image acquisition setup similar to the one described in \cite{follmann2018d2s} and attempt to automatically generate the pixel-wise object annotations. A handful of training images is acquired for each object in D2S such that each view of the object is captured. The training images are kept very simple. Each contains only instances of a single class and the instances do not touch each other. During the acquisition process, only the category of the object on the turntable has to be set manually. This does not result in additional work, as one has to collect the classes that have already been captured, anyway. Together with the simple acquisition setup, these restrictions allow to generate the pixel-precise annotations of the objects automatically. We present different approaches, one with background subtraction and the other based on salient object detection.
    \subsection{Acquisition}
\begin{figure}[t]
  \centering
  \includegraphics[width=0.35\textwidth]{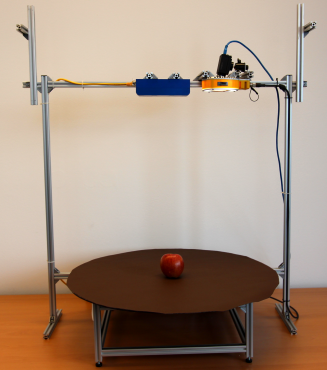}
  \hspace{2cm}
  \includegraphics[width=0.35\textwidth]{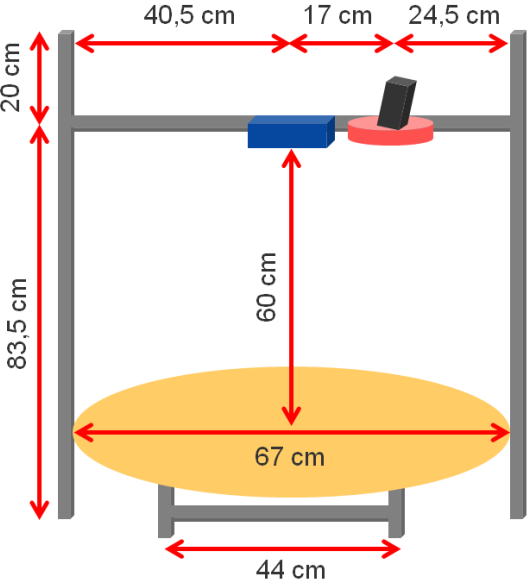}
  \caption{The \emph{D2S} image acquisition setup.
    Each scene was rotated ten times using a turntable. For each rotation, three images are acquired with  different illuminations.}
  \label{fig:aufbau}
\end{figure}

    To be able to reduce the label and acquisition effort to a bare minimum, the image acquisition setup is constructed in a very basic manner. A high-resolution industrial RGB camera with $1920 \times 1440$ pixels is mounted above a turntable. The turntable allows to acquire multiple views of each scene without any manual interaction. To increase the perspective variation, the camera is mounted off-center with respect to the rotation center of the turntable. Additionally, a stereo camera that works with projected texture is fixed centrally above the turntable. In \secref{sec:data_augmentation}, we show how the depth images may be used to extend the capabilities of data augmentation. The setup used for the image acquisition and its dimensions are depicted in \figref{fig:aufbau}.

\begin{figure}
 \centering
 \includegraphics[width=0.235\textwidth]{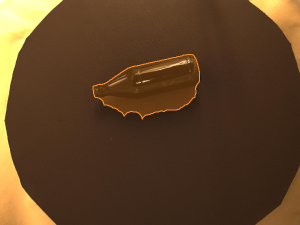};
 \includegraphics[width=0.235\textwidth]{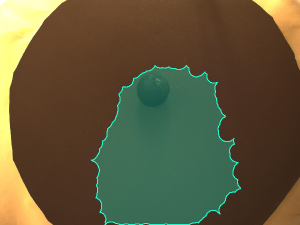};
 \includegraphics[width=0.235\textwidth]{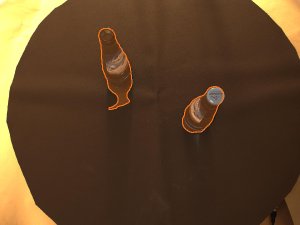};
 \includegraphics[width=0.235\textwidth]{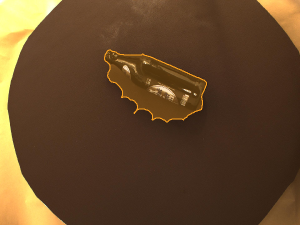};
  \includegraphics[width=0.235\textwidth]{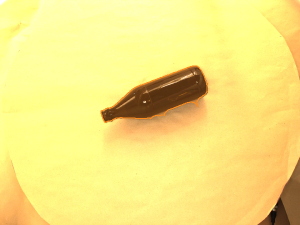};
 \includegraphics[width=0.235\textwidth]{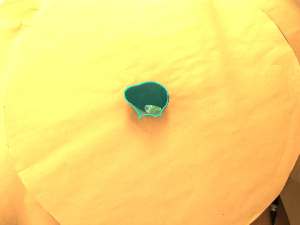};
 \includegraphics[width=0.235\textwidth]{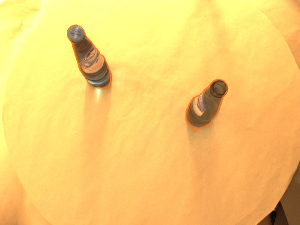};
 \includegraphics[width=0.235\textwidth]{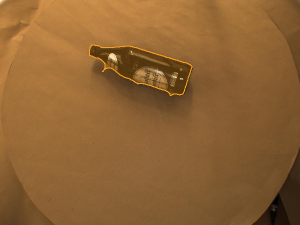}
 
 \caption{Examples where the light background used for \texttt{weakly\_cleaned} significantly improves the automatic segmentations over those from the dark background used for \texttt{weakly}.}
 \label{fig:cleaned}
 \end{figure}

To make the automatic label generation as simple and robust as possible, the background of the turntable is set to a plain colored brown surface. In an initial step, we keep the background color fixed for every training image. In a next step, we further use a lighter brown background to improve the automatic segmentation of dark or transparent objects such as avocados or bottles. The datasets are denoted with the prefix \texttt{weakly} and  \texttt{weakly cleaned}, respectively. A few example classes where the lighter background significantly improved the automatic labels are displayed in \figref{fig:cleaned}.
%
%
%
%

Note that, although the acquisition setup is very similar to that of the original D2S setup, there is a domain shift between the new training images and original D2S images. In particular, the camera pose relative to the turntable is not the same and the background and the lighting slightly differs. Maybe the most significant differences is, however, that some of the captured objects are different from those in original D2S. For example the vegetables and fruit categories have a slightly different appearance and some packaging, e.g., for the classes \texttt{clementine} or \texttt{apple\_braeburn\_bundle} are not the same as in D2S. The two classes \texttt{oranges} and \texttt{pasta\_reggia\_fusilli} were not available for purchase anymore and therefore, the respective D2S training scenes (without labels) were used.

    \subsection{Background Subtraction}
    \label{sec:background_substraction}
\begin{figure}
 \centering
 \begin{tikzpicture}
 [
 	 box/.style={rectangle,draw=black,thick, minimum size=1cm},
 ]
 
 \node[inner sep=0pt] (target) [label=270:Image] at (-5,0)
 {\includegraphics[width=0.18\textwidth]{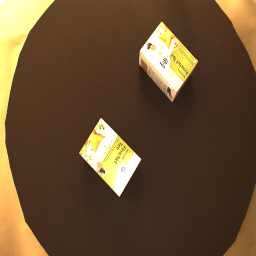}};
 
 \node[inner sep=0pt] (a) [] at (-3.4,0)  {$\ominus$};
 
 \node[inner sep=0pt] (target) [label=270:Background] at (-1.8,0)
  {\includegraphics[width=0.18\textwidth]{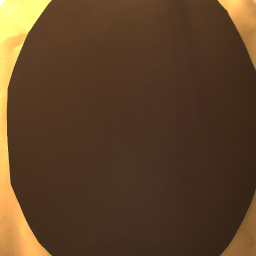}};
  
	\node[inner sep=0pt] (a) [] at (-0.1,0)  {$=$};
  
  \node[inner sep=0pt] (target) [label=270:Difference] at (1.6,0)
   {\includegraphics[width=0.18\textwidth]{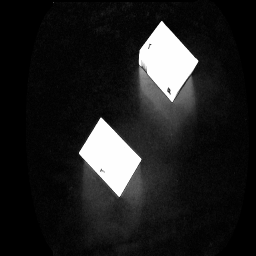}};
  
 \path[->] (3.0,0) edge[bend left] node[above] {} (3.6,0); 
 
 \node[inner sep=0pt] (score) [label=270:Annotations] at (5.0,0)
 {
 \includegraphics[width=0.18\textwidth]{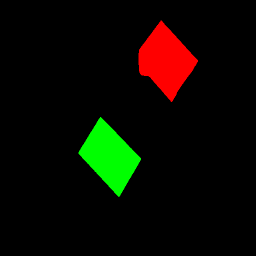}
 };
   
 \end{tikzpicture}
 \caption{Schematic overview of the weakly supervised ground truth instance segmentation generation.}
 \label{fig:flowchart}
 \end{figure}
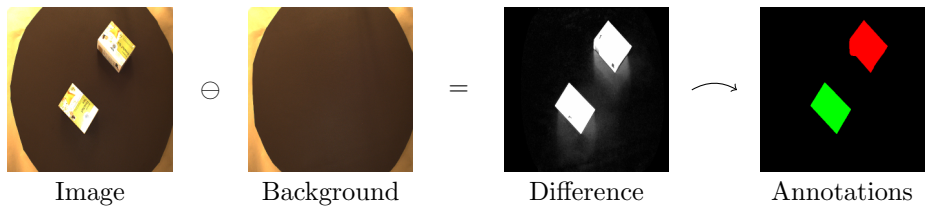

    We utilize the simple setting of the training images and automatically generate the segmentations by background subtraction. To account for changing illumination of the surrounding environment, an individual background image was acquired for each training scene. By subtracting the background image from each image, a foreground region can be generated automatically with an adaptive binary threshold \cite{otsu_1979_threshold}. Depending on the object and its attributes, we either use the V channel from the HSV color space, or the summed absolute difference of each of the RGB channels. The results for both color-spaces can be computed with negligible cost. Therefore, they can already be shown during the acquisition process and the user can choose the better region online. To ensure the object is not split into multiple small parts, we perform morphological closing with a circular structuring element on the foreground region. The instance segmentations can then be computed as the connected components of the foreground. The automatic segmentation method assumes that the objects are not touching or occluding each other, and generally works for images with an arbitrary number of objects of the same category. A schematic overview of the weakly supervised region generation is shown in \figref{fig:flowchart}. The resulting training set is denoted as \texttt{weakly} or \texttt{weakly\_cleaned} when using a lighter background for dark objects.
    
    \subsection{Saliency Detection}
%
%
As an alternative to the algorithmically simple background subtraction, the characteristics of the training images also invite to use saliency detection methods to identify the instances. Currently, the best methods are based on deep learning and require fine-tuning to the target domain \cite{borji2015salient,li2015inner}. Hence, they require manually labeling at least for a subset of the data. A more generic approach is that of the \emph{Saliency Tree} \cite{liu2014saliency}. It is constructed in a multi-step process. In a first step, the image is simplified into a set of primitive regions. The partitions are merged into a saliency tree based on their saliency measure. The tree is then traversed and the salient regions are identified. The salient region generation requires no fine-tuning to the target domain and achieves top ranks in recent benchmarks \cite{borji2015salient,li2015inner}.

\begin{figure}
 \centering
 \includegraphics[width=0.23\textwidth]{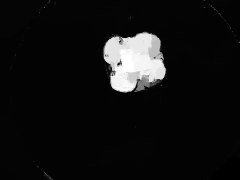}\;
 \includegraphics[width=0.23\textwidth]{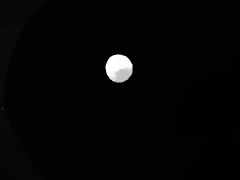}\;
 \includegraphics[width=0.23\textwidth]{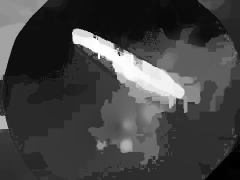}\;
 \includegraphics[width=0.23\textwidth]{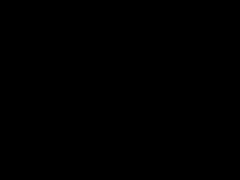}
 \includegraphics[width=0.23\textwidth]{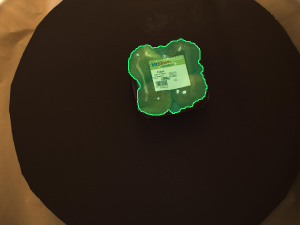}\;
 \includegraphics[width=0.23\textwidth]{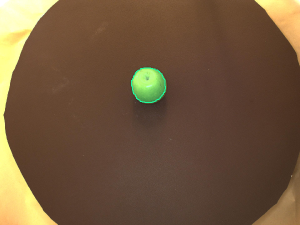}\;
 \includegraphics[width=0.23\textwidth]{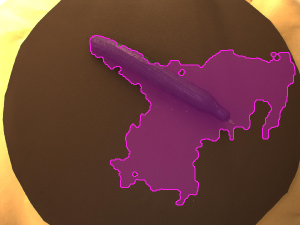}\;
 \includegraphics[width=0.23\textwidth]{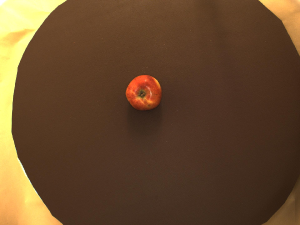}
 \includegraphics[width=0.23\textwidth]{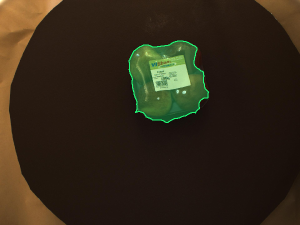}\;
 \includegraphics[width=0.23\textwidth]{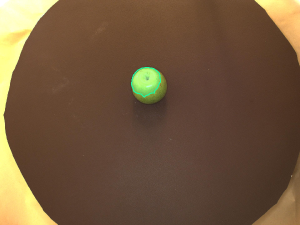}\;
 \includegraphics[width=0.23\textwidth]{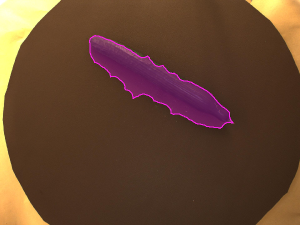}\;
 \includegraphics[width=0.23\textwidth]{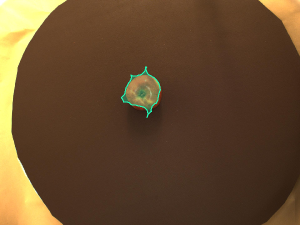}
 \caption{The automatic labels for saliency detection (second row) and background subtraction (third row) are displayed. In general, the saliency detection and the background subtraction return similar results (first column). In rare cases, the saliency detection outperforms the background subtraction and returns more complete regions. In the third and fourth column some typical failure cases of the saliency detection scheme are displayed. Either it fails completely (fourth column), or it is hard to find a reasonable threshold (third column).}
 \label{fig:segmentationexamples}
 \end{figure}

    We use the Saliency Tree to generate saliency images for each of the (cleaned) training images. The foreground region is then generated from the saliency image by a simple thresholding and an intersection with the region of the turntable. Also here, morphological closing and opening with a small circular structuring element is used to close small holes and smooth the boundary. Analogously to the background subtraction, the single instances are computed as the connected components of the foreground region. Qualitatively, we found that a threshold of 40 was a good compromise between too large and too small generated regions. For some rather small objects, using this threshold results in regions that almost fill the whole turntable. To prevent those artifacts, we iteratively increase the threshold by ten until the obtained total area of the regions is smaller than $30\%$ of the turntable area. However, even with this precaution, in some cases the obtained instances may be degenerated. A few examples and failure cases or both the background subtraction and the saliency detection are displayed in \figref{fig:segmentationexamples}. We denote the annotations obtained with the saliency detection method by \texttt{saliency cleaned}.

    \section{Data Augmentation} \label{sec:data_augmentation}
    
    One of the challenges of applying deep-learning based CNNs is the large amount of training data that is required to obtain competitive results. In the real-world applications discussed in this work, the acquisition and labeling of training data can be expensive as it requires many manual tasks. To mitigate this issue, we use data augmentation, where additional training data is created automatically based on a few manually acquired simple images. The images generated by the data augmentation model and simulate the variations and complexity that commonly occur when applying the trained network.
    
    To augment the training data, we randomly select between 3 and 15 objects from the training set, crop them out of the training image utilizing the generated annotation and paste them onto a background image similar to the one from D2S training images. The objects are placed at a random location with a random rotation. This generates complex scenes, where multiple objects of different instances may be overlapping each other. However, since this does not address all the difficulties within the D2S validation and test set, we also introduce two new augmentation techniques to specifically address these difficulties, namely {\it touching objects} and {\it reflections}. 
    
     \subsection{Touching Objects}
     
\begin{figure}
      \centering
      \includegraphics[width=0.235\textwidth]{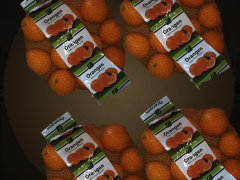}
      \includegraphics[width=0.235\textwidth]{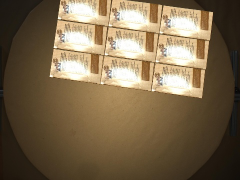}
      \includegraphics[width=0.235\textwidth]{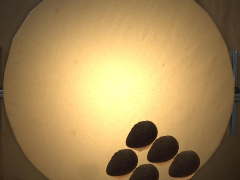}
      \includegraphics[width=0.235\textwidth]{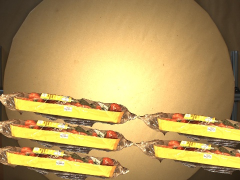}
      \caption{Examples for augmented reflections using a simulation of a spotlight.}
      \label{fig:neighboring}
    \end{figure}

      In the validation and test set of D2S there are many instances of the same class that touch each other. The existing instance segmentation schemes have difficulties to find the instance boundaries and often return unsatisfactory results. Often, if objects are close, the methods only predict a single instance or the instances extend into the neighboring object. Hence, we specifically augment the training set by generating new images where instances of the same class are very close to or even touching each other. We denote the respective dataset with the suffix \texttt{neighboring}. \figref{fig:neighboring} shows some examples of augmented touching objects.
    
    \subsection{Reflections}
\begin{figure}
 \centering
  \includegraphics[width=0.235\textwidth]{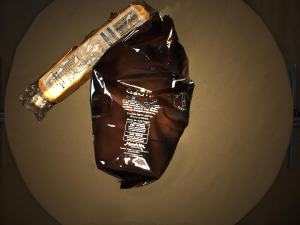}
  \includegraphics[width=0.235\textwidth]{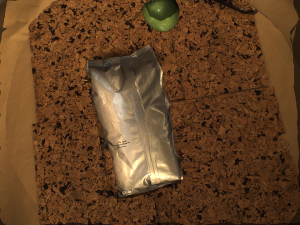}
  \includegraphics[width=0.235\textwidth]{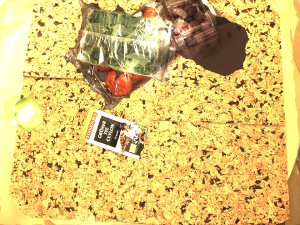}
  \includegraphics[width=0.235\textwidth]{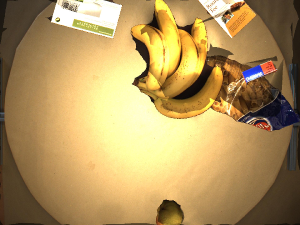}
 \caption{Examples for augmented reflections using a simulation of a spotlight.}
 \label{fig:reflections}
 \end{figure}

To create even more training data, we augment the original data by rendering artificial scenes under different lighting conditions. For this, we use the registered 3D sensor and RGB camera to build textured 3D models of the different object instances. Random subsets of those instances are then placed at random locations to create new, artificial scenes. Since we do not know the surface characteristics of the individual objects, we use a generic Phong shader \cite{phong1975illumination} with varying spotlight location and spotlight and ambient light intensity to simulate real-world lighting. We use this approach also because in real-world scenarios, lighting can vary drastically compared to the training setup. For example, different checkout counters can have different light placements, while others might be close to a window such that the time of day and the weather influence the local lighting conditions.  We denote the respective dataset with the suffix \texttt{reflections}. Example images of the \texttt{reflections} set are shown in \figref{fig:reflections}.

%
%
    \section{Experiments}
    
    All of the experiments are carried out on D2S \cite{follmann2018d2s}, as the datasets splits are explicitly designed for the use of data augmentation. In comparison to the validation and test sets, the complexity of the scenes in the training set are a lot lower in terms of object count, occlusions, and background variations. Moreover, the data augmentation techniques introduced in \secref{sec:data_augmentation} are well-suited for industrial setting, where the intra-class variations are mainly restricted to rigid and deformable transformations and background or lighting changes. D2S contains 60 object categories in 21.000 pixel-wise labeled images.

    We do not want to carry out a review of instance segmentation methods, but focus on the analysis of the weakly supervised setting and the different augmentations. Therefore, we use the competitive instance segmentation method Mask R-CNN \cite{he_2017_mask} for our experiments. We choose the original Detectron \cite{Detectron2018} implementation in order to make the results easy to reproduce.
    
    \paragraph{\textbf{Setup.}}

To speed up the training and evaluation process, we downsized D2S by a factor of 4, which results in an image resolution of $480\times360$ px. Note that we also adapted the image size parameters of Mask R-CNN accordingly, such that no scaling has to be carried out during training or evaluation.

All models were trained using two NVIDIA GTX1080Ti GPUs. We used the original hyperparameters from Detectron except for the following: a batch size of 5 images per GPU (resulting in 10 images per iteration) and a base learning rate of 0.02 (reduced to 0.002 after 12k iterations). We trained for 15000 iterations and initialized with weights pretrained on COCO.
    
    For the evaluation of our results, we used the \emph{D2S} validation set. As is common in instance segmentation, the mean average precision $[\%]$ averaged over IoU-thresholds from 0.5 to 0.95 in steps of 0.05 is used as performance measure.
    
    \paragraph{\textbf{Baseline.}}
    As a baseline for the weakly supervised setting, we used the high quality, manually generated annotations from the D2S training set. To better compare and separate the effect of the data augmentation from the effect of having better training data, we also augmented this high quality training set.
    Except for reflections (which requires depth information), the augmentation can be done analogously to the weakly supervised setting. Because the annotations fit almost perfectly, the object crops contain only a very small amount of background surroundings compared to the crops from the weak annotations. Therefore, one can expect the best results for the baseline.
    
    \paragraph{\textbf{Results.}}
    
    For all types of underlying annotations, \texttt{baseline}, \texttt{weakly}, \texttt{weakly cleaned} and \texttt{saliency}  \texttt{cleaned}, we made similar experiments. First, we trained the model only on the training images, denoted as \texttt{base}. Second, we augmented 2500, 5000, 10000, or 20000 images as described at the beginning of \secref{sec:data_augmentation} and added them to \texttt{base}, respectively (\texttt{augm}). Third, we augmented 2000 images both with touching objects (\texttt{neighboring}) or on a random background (\texttt{random} \texttt{background}). Additionally, for \texttt{weakly}  \texttt{cleaned} and \texttt{saliency}  \texttt{cleaned}, we generated 2000 images with \texttt{reflections} (on random background). The corresponding mAP values on the D2S validation set (in quartersize) are shown in \tabref{tbl:results}.
    
    \begin{table}
  \centering
  \caption{{\bf Results.} Ablation study for different ways of generating the annotations and different augmentations. * indicates the set that gave the best results in combination with specific augmentations. Abbreviations for augmentation types are as follows: \texttt{neighboring} (NB), \texttt{random background} (RB), \texttt{reflections} (RE).}
  \label{tbl:results}
  \footnotesize
  \def\arraystretch{1.2}
  \setlength{\tabcolsep}{8pt}
  \begin{tabular}{l|c|c|c|c}
    \hline
                   & \multirow{2}{*}{baseline} & \multirow{2}{*}{weakly} & weakly  & saliency \\
      Training Set &                           &                         & cleaned & cleaned \\
    \hline\hline
    \texttt{base}                   & 48.3 &  8.5 & 15.9 & 16.5 \\ \hline
	\texttt{base} + \texttt{augm 2500}  & 77.0 & 62.8 & 64.8 & 55.7 \\
    \texttt{base} + \texttt{augm 5000}* & 77.8 & 62.2 & 61.9 & 55.4 \\
    \texttt{base} + \texttt{augm 10000} & 78.4 & 65.0 & 63.0 & 55.8 \\
    \texttt{base} + \texttt{augm 20000} & 78.4 & 65.0 & 62.7 & 59.7 \\ \hline
    * + NB                              & 78.5 & 63.7 & 62.5 & 59.2 \\
    * + RB                              & 78.5 & 64.9 & 68.0 & 58.7 \\
    * + RE                              & -    & -    & 66.9 & 59.7 \\ \hline
    * + NB + RB                         & \textbf{80.1} & \textbf{66.8} & \textbf{68.9} & 60.2 \\
    * + NB + RB + RE                    & -    & -    & 68.5 & \textbf{61.9} \\
	\hline
  \end{tabular}
\end{table}
    
\begin{figure}
 \centering
 \begin{tabular}{c c c c}
 ground truth & base & + augm 5000 & + specific \\
 \includegraphics[width=0.235\textwidth]{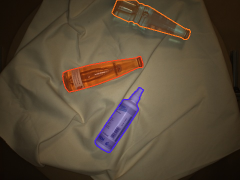} &
 \includegraphics[width=0.235\textwidth]{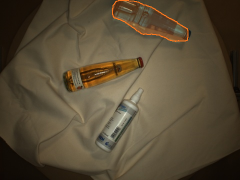} &
 \includegraphics[width=0.235\textwidth]{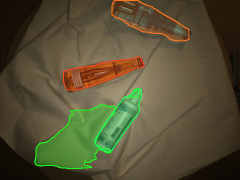} &
 \includegraphics[width=0.235\textwidth]{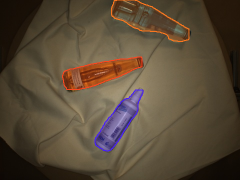} \\
 \includegraphics[width=0.235\textwidth]{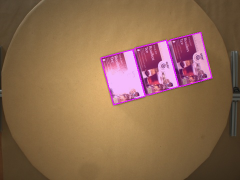} &
 \includegraphics[width=0.235\textwidth]{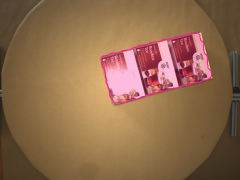} &
 \includegraphics[width=0.235\textwidth]{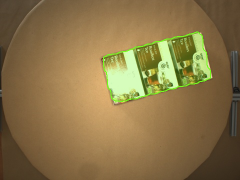} &
 \includegraphics[width=0.235\textwidth]{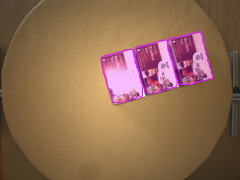} \\
 \includegraphics[width=0.235\textwidth]{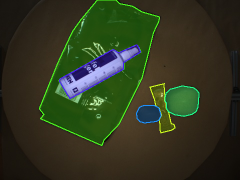} &
 \includegraphics[width=0.235\textwidth]{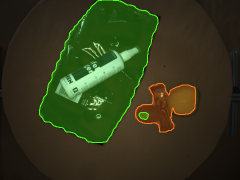} &
 \includegraphics[width=0.235\textwidth]{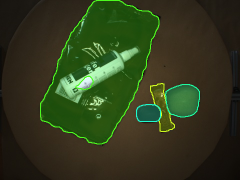} &
 \includegraphics[width=0.235\textwidth]{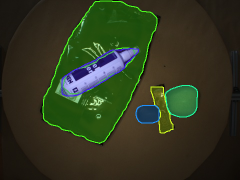} \\
  \end{tabular}
 \caption{{\bf Qualitative Results.} Improvements for \texttt{weakly\_cleaned} using specific augmentations: (\emph{top}) \texttt{random backgrounds}, (\emph{middle}) \texttt{neighboring} objects, (\emph{bottom}) \texttt{reflections}.}
 \label{fig:results}
 \end{figure}

    The images obtained for our weakly supervised training are significantly less complex than the D2S validation images; there are no touching or occluding objects and always only one category per image. This large domain shift results in a very poor performance of the models trained only on the \texttt{base} compared to the \texttt{baseline} (\cf row 1 of \tabref{tbl:results}). Normal augmentation strongly improves the results, e.g., from 8.5\%  to 65.0\% for \texttt{weakly}. Note that for the normal augmentation, the annotation quality seems to be less important, as \texttt{weakly cleaned} is on the same mAP-level as \texttt{weakly}. Only \texttt{saliency cleaned} performs significantly worse, probably due to some corrupt automatically generated annotations. The specific annotations types \texttt{neighboring} (NB), \texttt{random}  \texttt{background} (RB) and \texttt{reflections} (RE) further help to improve the result to 68.9\% for \texttt{weakly cleaned}, which is more than four times better than the \texttt{base}-result. NB, RB and RE are indeed complementary augmentation types as each of them consistently helps to improve upon \texttt{base + augm 5000}. In \figref{fig:results} some qualitative results are displayed. They show that using the specific augmentations indeed helps to improve on the typical failures cases that they address. Also note that the relative improvement using specific augmentations is higher in the weakly setting than for the \texttt{baseline} (\eg 7\% for \texttt{weakly cleaned} \vs 2.3\% for \texttt{baseline}).
    
     Usually with a higher number of training data the results of models with a high number of parameters are improved. However, we found that the best results are obtained if the specific augmentation sets of step three and four are added to \texttt{base + augm 5000}. A reason could be the domain-shift between D2S validation and the augmented images. For completeness, we show results for \texttt{augm 2500} and \texttt{augm 10000} in the supplementary material.
     
    \section{Conclusion}
We have presented a system that allows to train competitive instance segmentation methods with virtually no label effort. By acquiring very simple training images, we were able to automatically generate reasonable object annotations for the D2S dataset. To tackle the complex validation and test scenes, we propose to use different types of data augmentation to generate artificial scenes that mimic the expected validation and test sequences. We present new data augmentation ideas to help improve scenes where touching objects and changing illumination is a problem. The results indicate that the weakly supervised models yield a very good trade-off between annotation effort and performance. This paves the way for cost-effective implementations of semantic segmentation approaches by lifting the requirement of acquiring large amounts of training images.

Using imperfect annotations, we also found that increasing the number of augmented images does not always improve the result. We believe that reducing the domain-shift to the test set by generating more realistic augmentations is an open topic that could resolve this problem. Additionally, we found that data augmentation can be beneficial even if the number of labeled training images is already large.

\bibliographystyle{splncs03}
\bibliography{egbib}

\appendix
\section*{Appendix}
In \tabref{tbl:results_baseline}, \tabref{tbl:results_weakly} and \tabref{tbl:results_weakly_cleaned} we show the influence of augmenting a different amount of images and adding specific augmentations. Abbreviations for augmentation types are as follows: \texttt{neighboring} (NB), \texttt{random background} (RB), \texttt{reflections} (RE).

\begin{table}[ht]
  \begin{minipage}{.5\linewidth}
    \caption{{\bf Baseline results}}
    \label{tbl:results_baseline}
    \begin{center}
    \footnotesize
    \def\arraystretch{1.2}
    \setlength{\tabcolsep}{4pt}
    \begin{tabular}{l|c|c|c}
      \hline
      & \texttt{augm} & \texttt{augm} & \texttt{augm} \\
      Training Set & 2500 & 5000 & 10000 \\ \hline \hline
      \texttt{base}                   & 48.3 & 48.3 & 48.3 \\ \hline
      \texttt{base} + \texttt{augm}   & 77.0 & 77.8 & 78.4  \\ \hline
      + NB                          & 77.3 & 78.5 & 78.3 \\
      + RB                          & 78.2 & 78.5 & 79.1 \\ \hline
      + NB + RB                     & 79.3 & \textbf{80.1} & 79.9 \\
      \hline
    \end{tabular}
    \end{center}
  \end{minipage}
  \begin{minipage}{.5\linewidth}
    \caption{{\bf Weakly results}}
    \label{tbl:results_weakly}
    \begin{center}
    \footnotesize
    \def\arraystretch{1.2}
    \setlength{\tabcolsep}{4pt}
    \begin{tabular}{l|c|c|c}
      \hline
      & \texttt{augm} & \texttt{augm} & \texttt{augm} \\
      Training Set & 2500 & 5000 & 10000 \\ \hline \hline
      \texttt{base}                   & 8.5 & 8.5 & 8.5 \\ \hline
      \texttt{base} + \texttt{augm}   & 62.8 & 62.2 & 65.0  \\ \hline
      + NB                          & 64.0 & 63.7 & 65.8 \\
      + RB                          & 64.0 & 64.9 & 63.5 \\ \hline
      + NB + RB                     & 64.0 & \textbf{66.8} & 65.3 \\
      \hline
    \end{tabular}
  \end{center}
  \end{minipage}
\end{table}

\begin{table}[ht]
  \caption{{\bf Weakly cleaned results}}
  \label{tbl:results_weakly_cleaned}
  \begin{center}
  \footnotesize
  \def\arraystretch{1.2}
  \setlength{\tabcolsep}{8pt}
  \begin{tabular}{l|c|c|c}
    \hline
    & \texttt{augm} & \texttt{augm} & \texttt{augm} \\
    Training Set & 2500 & 5000 & 10000 \\ \hline \hline
    \texttt{base}                   & 15.9 & 15.9 & 15.9 \\ \hline
    \texttt{base} + \texttt{augm}   & 64.8 & 61.9 & 63.0  \\ \hline
    + NB                          & 65.0 & 62.5 & 64.7 \\
    + RB                          & 65.9 & 68.0 & 65.8 \\
    + RE                          & 68.1 & 66.9 & 65.3 \\ \hline
    + NB + RB                     & 65.9 & \textbf{68.9} & 66.9 \\
    + NB + RB + RE                & 68.4 & 68.5 & 66.9 \\
    \hline
  \end{tabular}
  \end{center}
\end{table}

\end{document}